\documentclass{article}

\usepackage{microtype}
\usepackage{graphicx}
\usepackage{subcaption}
\usepackage{booktabs} 

\usepackage{hyperref}
\usepackage{xcolor}
\usepackage{arydshln} 

\usepackage[preprint]{icml2026}

\usepackage{amsmath}
\usepackage{amssymb}
\usepackage{mathtools}
\usepackage{amsthm}
\usepackage{multirow}
\usepackage{multicol}
\usepackage{enumitem}
\usepackage{graphicx}

\usepackage[capitalize,noabbrev]{cleveref}

\theoremstyle{plain}

\theoremstyle{definition}

\theoremstyle{remark}

\usepackage[textsize=tiny]{todonotes}

\icmltitlerunning{D-ORCA: Dialogue-Centric Optimization for Robust Audio-Visual Captioning}

\begin{document}

\twocolumn[
  \icmltitle{D-ORCA: Dialogue-Centric Optimization for Robust Audio-Visual Captioning}



  \icmlsetsymbol{equal}{*}

  \begin{icmlauthorlist}
    \icmlauthor{Changli Tang}{thu}
    \icmlauthor{Tianyi Wang}{wxg}
    \icmlauthor{Fengyun Rao}{wxg}
    \icmlauthor{Jing LYU}{wxg}
    \icmlauthor{Chao Zhang}{thu}
  \end{icmlauthorlist}

  \icmlaffiliation{thu}{Tsinghua University}
  \icmlaffiliation{wxg}{WeChat Vision, Tencent Inc.}

  \icmlcorrespondingauthor{Chao Zhang}{cz277@tsinghua.edu.cn}

  \icmlkeywords{Machine Learning, ICML}

  \vskip 0.3in
]



\printAffiliationsAndNotice{}  

\begin{abstract}
Spoken dialogue is a primary source of information in videos; therefore, accurately identifying who spoke what and when is essential for deep video understanding. We introduce D-ORCA, a \textbf{d}ialogue-centric \textbf{o}mni-modal large language model optimized for \textbf{r}obust audio-visual \textbf{ca}ptioning. We further curate DVD, a large-scale, high-quality bilingual dataset comprising nearly 40,000 multi-party dialogue videos for training and 2000 videos for evaluation in English and Mandarin, addressing a critical gap in the open-source ecosystem. To ensure fine-grained captioning accuracy, we adopt group relative policy optimization with three novel reward functions that assess speaker attribution accuracy, global speech content accuracy, and sentence-level temporal boundary alignment. These rewards are derived from evaluation metrics widely used in speech processing and, to our knowledge, are applied for the first time as reinforcement learning objectives for audio-visual captioning. Extensive experiments demonstrate that D-ORCA substantially outperforms existing open-source models in speaker identification, speech recognition, and temporal grounding. Notably, despite having only 8 billion parameters, D-ORCA achieves performance competitive with Qwen3-Omni across several general-purpose audio-visual understanding benchmarks. Demos are available at \href{https://d-orca-llm.github.io/}{https://d-orca-llm.github.io/}. Our code, data, and checkpoints will be available at \href{https://github.com/WeChatCV/D-ORCA/}{https://github.com/WeChatCV/D-ORCA/}. 
\end{abstract}

\section{Introduction}
\begin{figure*}[ht]
\begin{center}
\centerline{\includegraphics[width=0.9\linewidth]{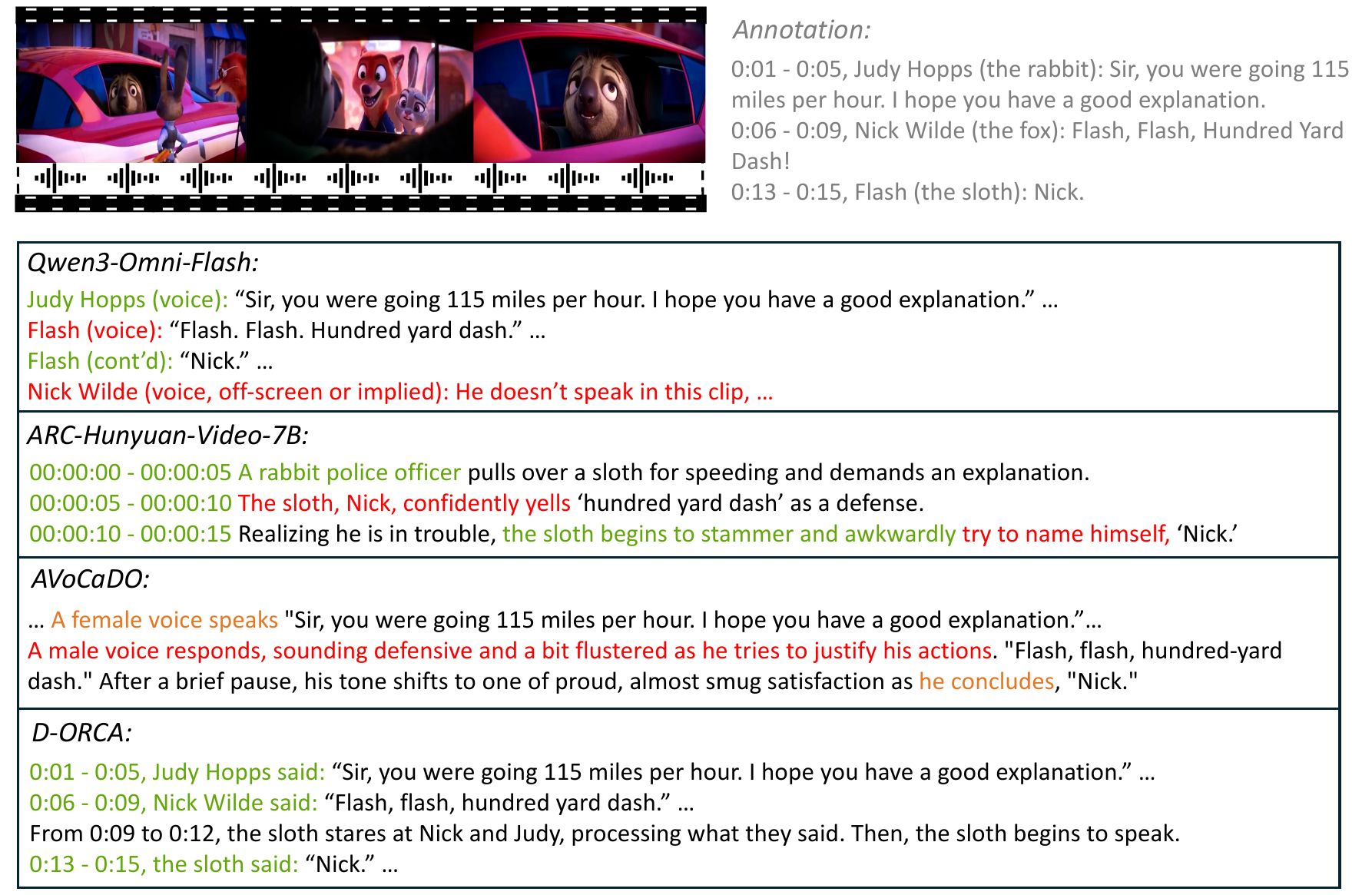}}
\caption{An example of video captions generated by different audio-visual LLMs for a dialogue-centric video clip. \textcolor[RGB]{116,165,51}{Green} highlights indicate correct speaker attribution, \textcolor{red}{red} indicates incorrect attribution, and \textcolor[RGB]{217,125,58}{orange} marks ambiguous or implicit speaker references. Existing open-source audio-visual LLMs struggle to accurately comprehend video dialogue, while our D-ORCA demonstrates more accurate and robust audio-visual understanding for dialogue-centric scenarios.}
\label{fig:model_case}
\vspace{-0.8cm}
\end{center}
\end{figure*}

Dialogue-centric content (\textit{e.g.}, films, television dramas, and everyday vlogs) constitutes a substantial portion of modern video data. In these scenarios, spoken dialogue drives the narrative, requiring audio-visual understanding systems to resolve not only what is said, but also who says it and when. Although recent visual-only \cite{zhang2024video, li2024llavaov, liu2024nvila, zhang2025videollama, chai2024auroracap, zhu2025internvl3, li2025improving, bai2025qwen2, bai2025qwen3vl, yao2024minicpm} and audio–visual \cite{gemini2.5, xu2025qwen2, tang2025video, ma2025omni, chen2025avocado, ge2025arc, sunvideo, ye2025omnivinci}  large language models (LLMs) offer promising end-to-end solutions, a critical gap remains: unlike humans, who naturally integrate auditory and visual cues (\textit{e.g.}, synchronizing voice with lip movements, or inferring off-screen speakers from static lips) for speaker attribution, current audio-visual LLMs struggle to achieve this level of cross-modal integration. As shown in Fig.~\ref{fig:model_case}, most existing open-source models fail to produce accurate, dialogue-aware captions, exposing a key limitation in current multimodal architectures.

The field also lacks benchmarks and evaluation metrics tailored to dialogue-centric audio–visual captioning. Standard captioning metrics such as BLEU~\cite{papineni2002bleu} and ROUGE-L~\cite{lin2004rouge} are inadequate for long-form dialogue, as they fail to capture semantic fidelity and speaker-specific accuracy. Recent metrics for detailed video captioning~\cite{chai2024auroracap, tang2025video, ma2025omni} emphasize overall content coverage but overlook the evaluation of synchronous audio–visual interactions essential to dialogue-driven scenarios. As a result, existing open-source models lack effective optimization signals for dialogue-centric understanding, often failing to correctly attribute speech to speakers or maintain coherent dialogue narratives.

To address these challenges, we first tackle the data scarcity problem by curating a large-scale video training set, DVD-Train, in dialogue-centric scenarios annotated with fine-grained \textbf{d}ialogue-centric \textbf{v}ideo \textbf{d}escription. Building on this foundation, we further construct DVD-Bench, a high-quality video benchmark for
dialogue-centric audio-visual understanding, which comprises diverse, human-annotated English and Chinese dialogue scenes to enable rigorous and systematic evaluation.
We argue that the core challenge of dialogue-centric video captioning lies in resolving the triplet of ``\textbf{when}, \textbf{who}, and \textbf{what} is said''. This requires accurately localizing utterances in time, associating them with the correct speakers, and precisely recognizing the spoken content. Despite recent advances, these abilities remain challenging for modern artificial intelligence (AI) models, and thus naturally define critical evaluation dimensions for coherent, dialogue-driven audio-visual understanding.
To explicitly optimize audio-visual LLMs for these capabilities, we propose a novel reinforcement learning (RL) framework that integrates group relative policy optimization (GRPO) \cite{shao2024deepseekmath} with three specialized reward functions, which guide the model toward fine-grained audio–visual alignment tailored to dialogue-centric scenarios:
\begin{itemize}[itemsep=0pt, leftmargin=*]
\item A speaker attribution accuracy reward that improves the binding between spoken content and the correct speaker;
\item A global speech content reward that enhances automatic speech recognition (ASR) accuracy;
\item A sentence-level temporal boundary reward that encourages precise utterance boundary localization.
\end{itemize}
Notably, these reward functions also serve as effective evaluation metrics for dialogue-centric audio-visual captioning. To ensure training stability and robustness, we introduce a pre–direct preference optimization (DPO) warm-up stage along with a reward balancing mechanism. Experiments show that the resulting model with 8 billion (B) parameters, D-ORCA, achieves state-of-the-art (SOTA) performance among open-source models for dialogue-centric audio-visual captioning in both English and Chinese. D-ORCA shows superior capability in speaker identification, precise temporal grounding, and robust speech recognition. Moreover, it exhibits strong generalization, achieving SOTA performance on general audio-visual question answering (QA) benchmarks among models of comparable scale.
Our main contributions are summarized as follows:
\begin{itemize}[itemsep=0pt, leftmargin=*]
\item We curate a large-scale, dialogue-rich bilingual video dataset DVD-Train as well as a high-quality benchmark DVD-Bench to advance research in dialogue-centric audio-visual understanding.
\item We propose a robust RL-based post-training framework with novel reward signals explicitly designed to resolve the challenges of ``{when}, {who}, and {what} is said''.
\item We present D-ORCA, which achieves SOTA results in dialogue-centric captioning while remaining highly competitive on broad audio-visual QA benchmarks.
\end{itemize}

\section{Background}
\subsection{Dialogue-centric Audio-Visual Understanding}

The core of dialogue-centric audio-visual understanding lies in resolving the triplet of ``when, who, and what is said''. Traditionally, the audio-only formulation of this problem has been investigated through speaker diarization pipelines, which typically involve voice activity detection to identify speech segments, speaker change point detection to partition them into speaker-homogeneous segments, and speaker embedding extraction followed by clustering for speaker identity assignment~\cite{park2022review}. To recover linguistic content, these modules are typically integrated into cascaded speaker-attributed ASR (SA-ASR) systems, where diarized segments are transcribed by downstream ASR models~\cite{raj2021integration, zheng2022tandem, cornell2023chime}. To reduce error propagation, recent work has shifted toward end-to-end SA-ASR. Existing approaches either assume a fixed speaker set and use multi-head architectures for speaker-wise transcription~\cite{kanda2020joint, lu2021streaming}, or handle open speaker scenarios via neural clustering or speaker indexing~\cite{zheng2025dncasr}.
More recently, the emergence of LLMs has introduced a new paradigm. Generative approaches such as SpeakerLM~\cite{yin2025speakerlm} and DiarizationLM~\cite{wang2024diarizationlm} have shown the potential of leveraging LLMs to perform end-to-end diarization and recognition within a unified sequence generation task.

In audio–visual settings, acoustic information alone is often insufficient, yet extending audio-centric pipelines to effectively incorporate visual cues remains challenging. Conventional approaches lack principled mechanisms to extract heterogeneous visual signals needed for accurate speaker attribution. Prior methods that rely on facial dynamics for diarization~\cite{xu2022ava, sharma2022using, wuerkaixi2022dyvise, he2022end, qiu2022visual, yang2023uncertainty, cheng2025multi, afouras2018deep} are brittle in complex scenarios (e.g., off-screen speakers) and lack holistic video understanding. The emergence of audio–visual LLMs that jointly process raw audio and video offers a promising alternative. A representative example is AVoCaDO~\cite{chen2025avocado}, which introduces dialogue-based rewards to optimize captioning based on speech content and speaker identity.


\subsection{Optimization for Audio-Visual Captioning}

Traditional video captioning benchmarks such as MSR-VTT~\cite{xu2016msr} and VATEX~\cite{wang2019vatex} primarily rely on short-form descriptions evaluated using n-gram–based metrics (\textit{e.g.}, BLEU, ROUGE-L), which are not suitable for assessing the fine-grained semantics of complex audio-visual scenes. As a result, recent work has increasingly focused on evaluation and optimization for detailed video captioning. In the visual domain, AuroraCap~\cite{chai2024auroracap} introduced a QA-based evaluation paradigm to measure content coverage. This idea was later extended to the audio-visual setting by UGC-VideoCaptioner~\cite{wu2025ugc}, which employs dual-modality question answering for evaluation and adopts GRPO with LLM-scored rewards during training. Similarly, AVoCaDO~\cite{chen2025avocado} leverages GRPO to refine captioning quality using checklist-based and dialogue-based rewards verified by external LLMs. video-SALMONN 2~\cite{tang2025video} achieves fine-grained assessment by decomposing video content into atomic events to quantify completeness versus hallucination, and guides model learning through multi-round DPO. 
Omni-Captioner~\cite{ma2025omni} emphasizes data scaling via an agentic framework that synthesizes dense annotations through iterative query–observation cycles.
Despite these advances, existing approaches for detailed audio-visual captioning largely emphasize the completeness of auditory and visual content, while overlooking the quality of joint audio-visual interactions, particularly those required for dialogue-centric captioning, such as speaker attribution, temporal alignment, and cross-modal consistency.

\section{Methods}
\begin{figure*}[ht]
\begin{center}
    \centerline{\includegraphics[width=\linewidth]{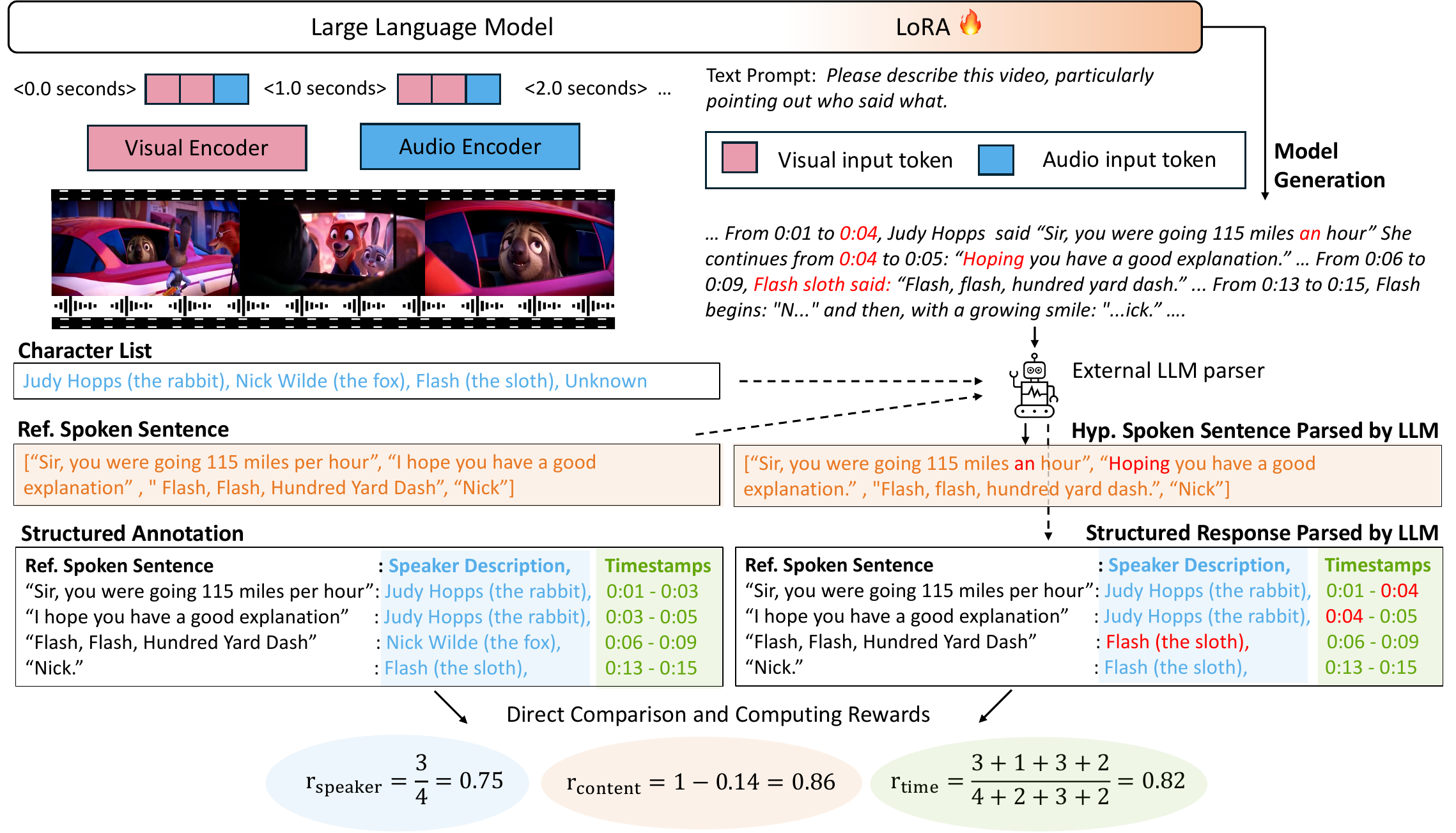}}
    \caption{Architecture and reward computation in D-ORCA. The model processes chronologically interleaved audio–visual tokens as input. During GRPO training, an external LLM parser is used to extract predicted ASR texts, and also identify the speaker and timestamp for each reference spoken sentence guided by a candidate character list. Based on these structured outputs, dialogue-centric rewards ($r_{\text{speaker}}$, $r_{\text{content}}$, $r_{\text{time}}$) are computed to optimize the accuracy of speaker attribution, speech content, and sentence-level temporal alignment.}
    \label{fig:structure}
    \vspace{-0.8cm}
\end{center}
\end{figure*}

\subsection{Model Architecture and Training Pipeline}
As illustrated in Fig.~\ref{fig:structure}, our architecture is designed to process paired audio–visual streams conditioned on user text prompts. The pipeline begins with parallel feature extraction from the input video. The visual stream is first temporally down-sampled into a sequence of frames and processed frame by frame using a pre-trained visual encoder. In parallel, the accompanying audio stream is processed by a pre-trained audio encoder. Both encoders are equipped with modality aligners that project the extracted visual and acoustic features into visual and audio input tokens compatible with the LLM. These modality-specific tokens are grouped according to their corresponding time intervals and interleaved to form a unified temporal sequence. Explicit textual timestamps are inserted as delimiters at the beginning of each temporal segment. The resulting multimodal token sequence is then concatenated with the user’s text prompt and fed into the LLM backbone. To improve training efficiency and parameter economy, we apply low-rank adaptation (LoRA)~\cite{hu2022lora} to the LLM.

As for the training pipeline, the model is built upon a strong visual LLM (\textit{e.g.}, Qwen3-VL). To endow this visual backbone with auditory perception and dialogue understanding capabilities, we first conduct audio modality alignment using large-scale audio datasets, followed by audio-visual supervised fine-tuning (SFT) on paired video-text data.
Specifically, in these stages, given the text prompt $\mathbf{q}$, the multimodal input sequence $\mathbf{x}$, and the target textual response $\mathbf{y}$, the training objective is to minimize the negative log-likelihood of the next-token prediction:
\vspace{-0.2cm}
\begin{equation}
    \mathcal{L}_{\text{SFT}}(\theta) = - \mathbb{E}_{{(\mathbf{x}, \mathbf{y}, \mathbf{q}})\sim \mathcal{D}}\left[\log \pi_\theta(\mathbf{y}|\mathbf{x}, \mathbf{q})\right],
\end{equation}
\vspace{-0.2cm}
where $\mathcal{D}$ and $\theta$ are the training set and model parameters.


Prior to optimizing dialogue-centric captioning with GRPO, we introduce a pre-DPO stage to mitigate the SFT model's repetition degeneration, which would otherwise destabilize reward calculation and hinder effective optimization. 
In the pre-DPO stage, $\forall\mathbf{x} \in \mathcal{D}$, two candidate captions are sampled from the SFT model. Inputs for which exactly one caption is complete and the other exhibits repetitive degeneration are retained to construct the pre-DPO training set $\mathcal{D}_{\text{DPO}}$. The complete caption and the repetitive caption are then assigned as the positive and negative samples, denoted by $\mathbf{y}_{\text{win}}$ and $\mathbf{y}_{\text{lose}}$, respectively. Under this construction, the training loss of the pre-DPO stage is defined as:


\begin{center}
    \vspace{-0.8cm}
    \small
    \begin{align}
        \nonumber &\mathcal{L}_\text{DPO}(\theta) =-\mathbb{E}_{(\mathbf{X},\mathbf{y}_\text{win},\mathbf{y}_\text{lose})\sim\mathcal{D}_\text{DPO}} \\ 
        & \left[\log \sigma \left(\beta \log \frac{\pi_{\theta}(\mathbf{y}_\text{win}\mid \mathbf{x})}{\pi_{\text{ref}}(\mathbf{y}_\text{win}\mid \mathbf{x})} - \beta \log \frac{\pi_{\theta}(\mathbf{y}_\text{lose}\mid\mathbf{x})}{\pi_{\text{ref}}(\mathbf{y}_\text{lose}\mid\mathbf{x})}\right)\right],
    \end{align}
    \vspace{-0.5cm}
    \label{equ:dpo}
\end{center}
where $\beta$ controls the deviation from the reference model $\pi_{\text{ref}}$, and $\sigma(\cdot)$ is the sigmoid function. 

Finally, GRPO is applied to optimize dialogue-centric video captioning, and a reward computation example is also shown in Fig~\ref{fig:structure}. Specifcally, for each input prompt $\mathbf{q}$, a group of $G$ outputs $\{\mathbf{y}_g\}_{g=1}^G$ from the old policy $\pi_{\theta_\text{old}}$ are generated, and we calculate rewards and normalize them within the group to obtain advantages $A_g$ for each output. The objective loss is formulated to maximize the expected reward while constraining policy updates, as shown below:
\vspace{-1cm}
\begin{center}
    \small
    \begin{align}\label{equ:grpo}
        \nonumber \mathcal{L}_{\text{GRPO}}(\theta) &= -\mathbb{E}_{(\mathbf{x},\mathbf{q})\sim \mathcal{D},\{\mathbf{y}_g\} \sim \pi_{\theta_\text{old}}} \\
        & \left[ \frac{1}{G} \sum_{g=1}^G \min \left( \rho_g A_g, \text{clip}(\rho_g, 1-\epsilon, 1+\epsilon) A_g \right) \right],
    \end{align}
\end{center}
where $\rho_g = {\pi_\theta(\mathbf{y}_g|\mathbf{x},\mathbf{q})}/{\pi_{\theta_\text{old}}(\mathbf{y}_g|\mathbf{x},\mathbf{q})}$ is the probability ratio, and $\epsilon$ is a hyper parameter of $\text{clip}(\cdot)$ that limits $1-\epsilon\leqslant\rho_g<1+\epsilon$. Note that the KL-divergence term in standard GRPO~\cite{shao2024deepseekmath} is not used.

\subsection{Dialogue-centric Reward Modeling}
\label{sec:reward_design}
To optimize the model’s ability to resolve when, who, and what is spoken in video dialogues, we propose a fine-grained reward modeling framework. Specifically, we design three complementary reward signals: \textit{Speaker Attribution Accuracy} ($r_{\text{speaker}}$), \textit{Global Speech Content Accuracy} ($r_{\text{content}}$), and \textit{Sentence-level Temporal Boundary Alignment} ($r_{\text{time}}$). An advanced external LLM is employed to parse the generated captions into a structured form, enabling direct comparison with the ground-truth annotations.

\noindent \textbf{Data Structure and Annotation.}
We formally represent the annotation for a video as a set of spoken sentence annotations $\mathcal{V} = {(s_i, p_i, \tau_i)}_{i=1}^N$, where $s_i$ denotes the reference transcript of the $i$-th utterance, $p_i$ specifies the speaker identity via a textual description, and $\tau_i = [t_{i,\text{start}}, t_{i,\text{end}}]$ defines the corresponding temporal interval. Here, $N$ denotes the total number of spoken sentences in the video. In addition, we maintain a global set of character descriptions $\mathcal{C}$ that covers all distinct roles appearing in the video. Note that some characters in $\mathcal{C}$ may not speak, and speaking characters may not appear within the video frames.

\noindent \textbf{Speaker Attribution Accuracy Reward $r_{\text{speaker}}$.}
This reward measures the model’s ability to correctly bind speech content to the corresponding speakers. We employ an external LLM to parse the generated caption and assign each ground-truth sentence $s_i$ to a predicted speaker $\hat{p}_i$ selected from a candidate set $\mathcal{C}$. To penalize ambiguous speaker attribution, we augment $\mathcal{C}$ with an ``Unknown'' option during training; sentences with unspecified or unresolvable speakers are assigned to this class.
The reward is computed as the proportion of sentences with correctly attributed speakers:
\begin{equation}
    r_{\text{speaker}} = \frac{1}{N} \sum_{i=1}^{N} \mathbb{I}(\hat{p}_i = p_i),
\end{equation}
where $\mathbb{I}(\cdot)$ is the indicator function.

\noindent \textbf{Global Speech Content Reward $r_{\text{content}}$.}
To ensure high-quality speech recognition within the captioning process, we evaluate transcription accuracy using word error rate (WER) for English and character error rate (CER) for Chinese. To reduce errors caused by cross-sentence word or character misalignment, we concatenate all ground-truth utterances $\mathbf{s}_i$ and the corresponding hypothesis utterances $\hat{s}_i$, extracted in chronological order from the generated captions by an external LLM, into unified transcripts $S$ and $\hat{S}$ for each video $\mathcal{V}$. The resulting content reward is then defined as:
\begin{equation}
    r_{\text{content}} = 
    \begin{cases} 
        1 - \text{WER}(S, \hat{S}), & \text{for English,} \\
        1 - \text{CER}(S, \hat{S}), & \text{for Chinese.}
    \end{cases}
\end{equation}

\noindent \textbf{Sentence-level Temporal Reward $r_{\text{time}}$.}
This reward encourages precise temporal localization of dialogue events. An external LLM extracts predicted start and end timestamps $[\hat{t}_{i,\text{start}}, \hat{t}_{i,\text{end}}]$ for each sentence $s_i$ from the generated caption, and rewards are computed based on intersection over union (IoU). We explicitly handle two unknown (Unk) cases during LLM parsing: \textbf{(i) Bilateral Unknown}, where no timestamp is produced and the predicted interval is set to $[\text{Unk}, \text{Unk}]$, yielding zero intersection and a union equal to the ground-truth duration $|\tau_i|$; and \textbf{(ii) Unilateral Unknown}, where only one boundary is generated (\textit{e.g.}, due to partial timestamps or sentence misalignment; see Appendix~\ref{app:time_unk}). In the latter case, instead of assigning zero IoU, we estimate the missing boundary by projecting the ground-truth duration $|\tau_i|$ (\textit{e.g.}, $[\hat{t}_{i,\text{start}}, \text{Unk}] \rightarrow [\hat{t}_{i,\text{start}}, \hat{t}_{i,\text{start}} + |\tau_i|]$).

After resolving Unk cases, we compute the intersection $I_i$ and union $U_i$ between predicted and ground-truth intervals. The reward is defined as the global IoU across all sentences:
\begin{equation}
    r_{\text{time}} = \frac{\sum_{i=1}^N I_i}{\sum_{i=1}^N U_i}.
\end{equation}

\noindent \textbf{Curriculum Learning Strategy.}
The final reward $r_{\text{total}}$ is computed as a weighted sum of the three components. To explicitly penalize repetition degeneration, we impose a strict length constraint: if the generated response contains more than $l_{\text{max}}$ tokens, $r_{\text{total}}$ is set to zero. In addition, we observe that applying temporal constraints too early in training can cause optimization collapse. To address this issue, we adopt a curriculum strategy in which the temporal reward is introduced only after the model has sufficiently stabilized. The total reward at training step $k$ is defined as:
\begin{equation}\label{equ:reward}
    \small
    r_{\text{total}} = 
    \begin{cases} 
    0 & \text{if } l > l_{\text{max}}, \\
    \lambda_1 r_{\text{speaker}} + \lambda_2 r_{\text{content}} & \text{else if } k < K_{\text{warmup}}, \\
    \lambda_1 r_{\text{speaker}} + \lambda_2 r_{\text{content}} + \lambda_3 r_{\text{time}} & \text{otherwise.}
    \end{cases}
\end{equation}
where $\lambda_1, \lambda_2, \lambda_3$ are balancing coefficients and $K_{\text{warmup}}$ is the threshold step for curriculum activation.

\section{Experimental Setup}
\subsection{Data Specifications}
For audio modality alignment, we train on LibriSpeech~\cite{panayotov2015librispeech}, CommonVoice~\cite{ardila2020common}, WavCaps~\cite{mei2024wavcaps}, AudioCaps~\cite{audiocaps}, Clotho~\cite{drossos2020clotho}, and a 3,500-hour (h) subset of WenetSpeech~\cite{zhang2022wenetspeech}.


For the subsequent audio-visual training stages, we construct a large-scale, high-quality video dataset, DVD-Train, explicitly focused on dialogue-rich scenarios. The videos are manually sourced from diverse genres, including movie clips, TV dramas, TV shows, and interviews, ensuring a wide coverage of speaking styles and acoustic environments. Specifically, we compile a total of about 40,000 videos, evenly split between English and Chinese, to serve as the foundation for the audio-visual SFT and pre-DPO. To support the reward-based optimization, we further curate a specific subset from this primary collection, consisting of approximately 8,500 videos in total, with a roughly equal distribution of English and Chinese content. Gemini-3.0-flash is used to annotate these videos into the structured format described in Sec.~\ref{sec:reward_design}.



For evaluation, we construct DVD-Bench, comprising 964 English and 1,014 Chinese videos with complex multi-party dialogue. Unlike the training data, DVD-Bench is carefully curated and annotated by human experts who provide precise speaker identities for each spoken sentence, establishing a reliable benchmark for evaluation. Additional details are provided in Appendix~\ref{app:data}.

\subsection{Model, training and evaluation specifications}
The visual encoder and the backbone LLM of D-ORCA are initialized from Qwen3-VL-8B \cite{bai2025qwen3vl}, while the audio encoder utilizes Whisper-large-v3 \cite{radford2023robust}. To map the Whisper features into the LLM's input space, a randomly initialized window-level Q-Former \cite{tang2024extending} is used as the audio modality aligner. This Q-Former is configured with a single query token and operates on a temporal window of 0.5 seconds, effectively compressing the audio input into a sequence of one token per 0.5 seconds. For visual processing, inputs are sampled at a frame rate of 2 frames per second and capped at a maximum of 128 frames, with a resolution limit of 176,400 pixels per frame. This results in a maximum number of visual tokens of about 10,000. The LoRA of the LLM backbone is with a rank of 128 and a scaling factor of 2.0.


In the initial audio modality alignment stage, only the audio modality aligner is trained while the LLM remains frozen. During subsequent audio–visual SFT, both modality aligners and LoRA parameters are unfrozen to enable joint multimodal learning. In the pre-DPO and GRPO stages, optimization is restricted to LoRA parameters.
LoRA merging is performed during pre-DPO following video-SALMONN~2~\cite{tang2025video}.
For GRPO, we set the group size to 8. Hyperparameters in Equ.~\eqref{equ:reward} are configured as $L_{\text{max}}{=}3072$, $K_{\text{warmup}}{=}500$, $\lambda_1{=}0.9$, $\lambda_2{=}0.1$, and $\lambda_3{=}0.1$, with a learning rate of $2\times10^{-5}$. Training cost details are reported in Table~\ref{tab:training_setup}.

\begin{table}[h]
\setlength{\tabcolsep}{3pt}
\centering
\caption{Training settings and resource allocation for each stage of D-ORCA. \#GPUs denotes the number of A800 GPUs used.}
\label{tab:training_setup}
\resizebox{\linewidth}{!}{
\begin{tabular}{lcccc}
\toprule
Stage & Batch Size & \#Updates & \#GPUs & Time Cost \\
\midrule
Audio Alignment & 1024 & 13,452 & 128 & 6h \\
Audio-Visual SFT & 32 & 10,520 & 32 & 8h \\
pre-DPO & 8 & 850 & 8 & 3h \\
GRPO & 8 & 1,000 & 64 & 75h \\
\bottomrule
\end{tabular}
}
\end{table}


To evaluate dialogue-centric audio–visual captioning, we mirror the reward design and measure speaker attribution accuracy, speech transcription fidelity in WER/CER, and sentence-level temporal accuracy in IoU. Consistent with GRPO training, we use Gemini-2.5-Flash as the external judge for caption parsing and evaluation.

\section{Experimental Results}
\subsection{Main Results}
\label{sec:main_results}
Table~\ref{tab:capres} presents a comprehensive quantitative comparison between D-ORCA and SOTA open-source omni-modal LLMs on our dialogue-centric audio-visual captioning benchmark. The results indicate that D-ORCA provides a holistic and structured understanding of the video narrative, demonstrating a distinct advantage in resolving the ``when, who, what is said'' of audio-visual interactions.


On the primary metric of speaker attribution accuracy, D-ORCA achieves strong performance in both English and Chinese settings. Notably, the base SFT model already performs competitively, surpassing most general-purpose omni-modal baselines and matching AVoCaDO, which empirically validates the high quality of our curated instruction-tuning dataset and highlights the importance of dialogue-centric data for reducing speaker attribution errors. Building on this strong foundation, our full training pipeline delivers substantial additional gains: the final D-ORCA model improves speaker attribution accuracy by 10\% over the SFT baseline and outperforms the previous SOTA, AVoCaDO, by approximately 8\%. These results underscore the effectiveness of our fine-grained reward modeling strategy in guiding the model to accurately interpret audio–visual context and attribute speech to the correct speakers.


Beyond speaker identification, D-ORCA also shows strong speech transcription capability. As shown by the WER/CER results in Table~\ref{tab:capres}, most existing audio-visual LLMs, such as ARC-Qwen-Video-Narrator, video-SALMONN~2+, Qwen2.5-Omni, and Qwen3-Omni, tend to generate coarse dialogue summaries, often omitting key utterances or failing to capture exact speech content, which leads to substantially higher error rates. In contrast, D-ORCA achieves consistently low WER/CER, indicating that its generated captions preserve precise and complete transcriptions of spoken dialogue. This capability is initially established during the SFT stage through verbatim-oriented supervision and is further enhanced by the global content reward during GRPO.


With respect to temporal precision, most existing audio-visual  LLMs disregard temporal grounding during caption generation, producing plain-text descriptions without explicit time alignment. In contrast, D-ORCA preserves fine-grained temporal awareness. Leveraging the sentence-level temporal reward, D-ORCA substantially improves its temporal localization performance, achieving IoU scores of 57\% and 38\% on English and Chinese videos, respectively. The lower IoU observed for Chinese videos likely reflects the greater difficulty of segmenting fragmented Mandarin speech. Given the short duration of typical utterances and the coarse 1-second resolution of our metric, these results indicate that D-ORCA attains relatively reliable temporal localization.

\begin{table*}[ht]
    \caption{Results of different omni-modal LLMs on our curated DVD-Bench, including English (En) subset and Chinese (Zh) subset. The speaker attribution accuracy (Acc), global speech WER/CER, and sentence-level temporal IoU are evaluated. ``-" denotes instances where valid metrics could not be computed (e.g., the model fails to generate timestamps or cannot perform Chinese ASR).}
    \centering
    \small
    \begin{tabular}{lcccccc}
    \toprule
    \multirow{2}{*}{\textbf{Model}} & \multicolumn{3}{c}{\textbf{DVD-Bench (En)}} & \multicolumn{3}{c}{\textbf{DVD-Bench (Zh)}} \\
    \cmidrule(lr){2-4} \cmidrule(lr){5-7} & Acc$\%\uparrow$ & WER$\%\downarrow$ & IoU$\%\uparrow$ & Acc$\%\uparrow$ &  CER$\%\downarrow$ & IoU$\%\uparrow$ \\
    \midrule
    ARC-Qwen-Video-Narrator (7B) \cite{ge2025arc} & 66.4 & 65.0 & 23.0 & 63.2 & 53.6 & 10.1 \\
    Qwen2.5-Omni (7B) \cite{xu2025qwen2} & 62.7 & 83.6 & - & 55.7 & 69.4 & - \\
    video-SALMONN 2+ (7B) \cite{tang2025video} & 66.6 & 94.0 & - & 59.9 & - & - \\
    AVoCaDO (7B) \cite{chen2025avocado} & 72.9 & 17.9 & - & 69.3 & - & - \\
    Qwen3-Omni-Instruct (30B-A3B) \cite{xu2025qwen3omni} & 67.8 & 91.3 & - & 63.5 & 60.6 & - \\
    \midrule
    Ours-SFT Model (8B) & 71.2 & 29.8 & 31.8 & 69.6 & 30.3 & 24.7 \\
    D-ORCA (8B) & \textbf{81.1} & \textbf{16.6} & \textbf{57.1} & \textbf{78.0} & \textbf{17.5} & \textbf{37.8} \\
    \bottomrule
    \end{tabular}
    \vspace{-0.5cm}
    \label{tab:capres}
\end{table*}

Moreover, D-ORCA exhibits robust generalization beyond dialogue-centric tasks. As shown in Table~\ref{tab:qares}, D-ORCA only uses a maximum visual input of 128 frames (about 10,000 tokens) to deliver superior performance on comprehensive benchmarks such as Video-MME \cite{fu2025video}, WorldSense \cite{hong2025worldsense}, AVUT \cite{yang2025audio}, Video-Holmes \cite{cheng2025video}, DailyOmni \cite{zhou2025daily}, and AV-SpeakerBench \cite{nguyen2025see}, which surpasses strong baselines of similar size and remains highly competitive even when compared with substantially larger models like Qwen3-Omni.

\begin{table*}[ht]
    \caption{Results of different omni-modal LLMs on general audio-visual understanding benchmarks.}
    \centering
    \resizebox{\textwidth}{!}{
    \begin{tabular}{lcccccc}
    \toprule
    \textbf{Model} & \textbf{Video-MME} & \textbf{WorldSense} & \textbf{AVUT} & \textbf{Video-Holmes} & \textbf{DailyOmni} & \textbf{AV-SpeakerBench} \\
    \midrule
    
    OmniVinci (7B) & 68.6 & 48.2 & - & - & 66.5 & - \\
    ARC-Qwen-Video-Narrator (7B) & 62.4 & 45.1 & - & 43.2 & 63.2 & 40.2 \\
    Qwen2.5-Omni (7B) & 64.3 & 47.8 & 66.3 & 43.7 & 62.7 & 42.3 \\
    video-SALMONN 2+ (7B) & \textbf{73.4} & 50.9 & 69.5 & 46.9 & 71.8 & 51.6 \\
    AVoCaDO (7B) & 65.9 & 49.9 & 70.0 & 47.2 & 69.8 & 51.6 \\
    Qwen3-Omni-Instruct (30B-A3B) & 70.5 & \textbf{54.0} & 72.0 & \textbf{54.1} & 69.8 & 54.1 \\
    \midrule
    D-ORCA (8B) & 72.9 & 53.7 & \textbf{76.1} & 48.5 & \textbf{78.5} & \textbf{55.0} \\
    \bottomrule
    \end{tabular}
    }
    \vspace{-0.4cm}
    \label{tab:qares}
\end{table*}

\vspace{-0.2cm}
\subsection{Impact of pre-DPO}
\vspace{-0.2cm}
Since the audio encoder is not native to the visual LLM backbone, the SFT model suffers from severe repetition degeneration when processing complex audio-visual inputs. As shown in Table~\ref{tab:predpo}, the SFT baseline exhibits a repetition rate of 21.8\%. This pathology critically destabilizes GRPO: frequent length violations result in zero rewards for a certain portion of samples, causing gradient sparsity that hinders effective optimization.

To validate the necessity of pre-DPO, we conducted a comparative ablation on English scenarios. We compared applying GRPO directly to the SFT model (SFT $\rightarrow$ GRPO) versus our proposed pipeline (SFT $\rightarrow$ pre-DPO $\rightarrow$ GRPO). To isolate the impact, we excluded the temporal reward for this experiment ($\lambda_1=0.9, \lambda_2=0.1, \lambda_3=0$) and trained both settings for 200 steps.
The results indicate that direct GRPO struggles to correct the decoding pathology, retaining a notable repetition rate of 4.3\% and achieving suboptimal performance. In contrast, pre-DPO effectively regularizes the output distribution, reducing to reptition ratio to 0.7\%. This allows the subsequent GRPO stage to focus entirely on fine-grained dialogue refinement, resulting in better speaker attribution and speech fidelity.

\begin{table}[ht]
    \caption{Experiment on the impact of pre-DPO. Results are reported on the English subset of DVD-Bench. ``Rep\%'' denotes the repetition ratio of generated captions, and "Acc\%" denotes the speaker attribution accuracy.}
    \vspace{-0.1cm}
    \centering
    \small
    \resizebox{\linewidth}{!}{
    \begin{tabular}{lccc}
    \toprule
    \multirow{2}{*}{\textbf{Method}} & \multicolumn{3}{c}{\textbf{DVD-Bench (En)}} \\
    \cmidrule(lr){2-4} & Rep$\%\downarrow$ & Acc$\%\uparrow$ & WER$\%\downarrow$ \\
    \midrule
    SFT & 21.8 & 71.2 & 29.8\\
    SFT $\rightarrow$ GRPO & 4.3 & 78.9 & 20.4\\
    \midrule
    SFT $\rightarrow$ pre-DPO & \textbf{0.7} & 71.9 & 18.7 \\
    SFT $\rightarrow$ pre-DPO $\rightarrow$ GRPO & 0.8 & \textbf{79.4} & \textbf{18.5} \\
    \bottomrule
    \end{tabular}
    }
    \label{tab:predpo}
\end{table}

\vspace{-0.3cm}
\subsection{Ablation on Reward Balancing in GRPO}
\vspace{-0.2cm}
Optimizing for dialogue-centric video understanding requires a delicate balance between multiple objectives: identifying speakers, transcribing content, and localizing speech. We conducted ablation studies on English scenarios to determine the optimal configuration of reward coefficients. All experiments were conducted for 1,000 steps, corresponding to about one epoch. Results are summarized in Table~\ref{tab:balance}.

\begin{table}[ht]
    \centering
    \caption{Ablation study of reward balancing strategies. Evaluation is performed on the English subset of DVD-Bench. ``Collapse'' indicates model degeneration where metrics cannot be computed.}
    \vspace{-0.1cm}
    \setlength{\tabcolsep}{2pt}
    \small
    \resizebox{\linewidth}{!}{
    \begin{tabular}{lc ccc}
    \toprule
    \multicolumn{2}{c}{\multirow{2}{*}{\textbf{Settings}}} & \multicolumn{3}{c}{\textbf{DVD-Bench (En)}} \\
    \cmidrule(lr){3-5} 
     & & Acc$\%\uparrow$ & WER$\%\downarrow$ & IoU$\%\uparrow$ \\
    \midrule
    \multicolumn{2}{l}{pre-DPO Model} & 71.9 & 18.7 & 35.5 \\
    \midrule
    \multicolumn{2}{l}{\textbf{GRPO Variants} ($\lambda_1:\lambda_2:\lambda_3$)} & & & \\
    \cmidrule{1-2}
    $1:0:0$ & w/o $K_{\text{warmup}}$ & 78.2 & 27.8 & - \\
    $9:1:0$ & w/o $K_{\text{warmup}}$ & 80.5 & 16.5 & - \\
    $9:1:1$ & w/o $K_{\text{warmup}}$ & \multicolumn{3}{c}{\textit{Collapse at early stage ($\sim$200 steps)}} \\
    $9:1:0.1$ & w/o $K_{\text{warmup}}$& \multicolumn{3}{c}{\textit{Collapse at mid stage ($\sim$600 steps)}} \\
    $9:1:1$ & $K_{\text{warmup}}=500$ & 79.8 & 16.8 & 55.1 \\
    \bottomrule
    \end{tabular}
    }
    \label{tab:balance}
    \vspace{-0.5cm}
\end{table}

Given that the pre-DPO model already possesses a reasonable ability to transcribe speech, the core challenge lies in correctly binding this speech to its speaker. Consequently, we identify the speaker attribution reward $r_{\text{speaker}}$ as the primary driver of optimization. 
However, our experiments reveal that optimizing exclusively for speaker attribution ($\lambda_1$:$\lambda_2$:$\lambda_3$=1:0:0) over-optimizes this single objective, leading to speech content recognition deteriorating significantly. The WER increases by about 9\% compared with the pre-DPO model. In addition, we also observe that the generation becomes unfluent and incoherent, suggesting the general linguistic capability of the model is impaired. Incorporating a small fraction of the content reward ($\lambda_1$:$\lambda_2$:$\lambda_3$=9:1:0) acts as a stable regularizer, maintaining a low WER of 16.5\% while achieving a high speaker accuracy of 80.5\%.


Incorporating temporal localization via $r_{\text{time}}$ introduces additional optimization challenges. When the temporal reward is applied from the start of training, we observe training collapse at early ($\sim$200 steps) or intermediate ($\sim$600 steps) stages, even under balanced ($\lambda_1:\lambda_2:\lambda_3 = 9:1:1$) or reduced ($9:1:0.1$) weighting schemes. This instability likely stems from the fact that the temporal reward not only requires precise localization but also implicitly constrains the generation format. When the model is still learning the semantic associations of who and what is spoken, the additional when constraint substantially enlarges the optimization space, encouraging degenerate strategies that exploit anomalous output formats to ``hack'' the temporal reward, ultimately leading to training collapse.

To mitigate this, we employ a curriculum strategy and set $K_{\text{warmup}}=500$, which is halfway through the epoch. This sequential approach introduces $r_{\text{time}}$ only after the model's basic dialogue-centric learning stabilizes, and effectively prevents reward hacking and ensures convergence, boosting the temporal IoU to 55.1\% without compromising speaker accuracy and speech recognition.


\vspace{-0.2cm}
\subsection{Robustness of LLM-based Evaluation Metrics}
\vspace{-0.2cm}
Since traditional n-gram metrics are inadequate for assessing the semantic granularity of detailed video captions, we leverage advanced external LLMs, such as Gemini-2.5-flash, to parse our dialogue-centric results for evaluation. Generally, utilizing LLMs as judges introduces potential risks, including stochasticity, sensitivity to prompt phrasing, and inherent biases toward certain output styles. However, in our framework, we mitigate these risks by decomposing the complex evaluation of dialogue-centric captioning into three atomic, objective extraction tasks: ``when, who, and what is said.'' This decomposition transforms the evaluation from a subjective scoring task into a rigorous information extraction problem, which is well within the capabilities of modern LLMs.

To validate the reliability of our metrics, we compared the evaluation results on our English benchmark using three distinct powerful LLMs as parsers: Gemini-2.5-flash, Qwen3-235B-A22B, and GPT-4.1-mini. As shown in Table~\ref{tab:extllm}, the results for speaker attribution accuracy and speech WER are consistent across all LLMs, since extracting spoken content and identifying the corresponding speaker constitutes a fundamental reading comprehension task that is straightforward for advanced LLMs. Evaluating temporal IoU for each speech sentence introduces some variability. This likely stems from the fact that different LLMs exhibit distinct preferences when handling edge cases, such as partial timestamps (\textit{e.g.}, a start time without an end time) or mismatches between the model's sentence segmentation and the ground truth.
However, the relative performance trends remain identical. D-ORCA consistently outperforms the pre-DPO baseline by a significant margin regardless of the judge employed. This cross-model consistency confirms that our proposed metric design is robust and provides a fair assessment of dialogue-centric audio-visual understanding.

\begin{table}[ht]
    \setlength{\tabcolsep}{2pt}
    \caption{Robustness analysis of evaluation metrics using different external LLMs as parsers. Models are evaluated on the English subset of DVD-Bench.}
    \vspace{-0.1cm}
    \centering
    \resizebox{\linewidth}{!}{
    \begin{tabular}{lcccc}
    \toprule
    \multirow{2}{*}{\textbf{Model}} & \multirow{2}{*}{\textbf{Parser LLM}} & \multicolumn{3}{c}{\textbf{DVD-Bench (En)}} \\
    \cmidrule(lr){3-5} & & Acc$\%\uparrow$ & WER$\%\downarrow$ & IoU$\%\uparrow$ \\
    \midrule
    \multirow{3}{*}{pre-DPO Model} & Gemini-2.5-flash & 71.9 & 18.7 & 35.5 \\
    & Qwen3-235B-A22B & 67.8 & 17.9 & 25.2 \\
    & GPT-4.1-mini & 71.6 & 17.6 & 31.3\\
    \midrule
    \multirow{3}{*}{D-ORCA} & Gemini-2.5-flash & 81.1 & 16.6 & 57.1 \\
    & Qwen3-235B-A22B & 78.0 & 16.6 & 47.9 \\
    & GPT-4.1-mini & 79.9 & 16.2 & 41.4\\
    \bottomrule
    \end{tabular}
    }
    \vspace{-0.4cm}
    \label{tab:extllm}
\end{table}

\vspace{-0.2cm}
\section{Conclusion}
\vspace{-0.2cm}
In this paper, we introduced D-ORCA, a dialogue-centric omni-modal LLM designed to resolve the ``who, what, and when'' of audio-visual narratives through a specialized reinforcement learning framework. To this end, we proposed three novel reward signals tailored for speaker attribution accuracy, speech recognition robustness, and temporal grounding, which simultaneously serve as effective evaluation metrics for dialogue-centric video captioning. By curating the high-quality dialogue-rich bilingual DVD-Train dataset and implementing a robust training pipeline, D-ORCA achieves precise grounding of speech to characters and timestamps. Experiments demonstrate that D-ORCA not only establishes superior performance on our curated DVD-Bench but also maintains competitive capabilities across general audio-visual understanding benchmarks. We believe that understanding character dialogue is fundamental to holistic audio-visual comprehension, and D-ORCA marks a significant step forward toward this goal.

\newpage
\section*{Impact Statement}
This paper presents work whose goal is to advance the field of Machine
Learning. There are many potential societal consequences of our work, none
which we feel must be specifically highlighted here.





\bibliography{example_paper}
\bibliographystyle{icml2026}

\newpage
\appendix
\onecolumn

\section{Details on Parsing Unilateral Unknown Timestamps}
\label{app:time_unk}

Extracting precise temporal intervals for spoken sentences from model-generated captions is not always straightforward due to the variability in generation styles. While an ideal output contains explicit start and end timestamps for every sentence, the model often produces ambiguous temporal cues. To handle this, we define two specific scenarios where the external LLM parser is required to generate a ``Unilateral Unknown'' interval (i.e., $[\hat{t}_{start}, \text{Unk}]$ or $[\text{Unk}, \hat{t}_{end}]$).

\noindent \textbf{Scenario 1: Discrete Time Points.}
The first scenario occurs when the model provides a single timestamp marking the onset of a sentence, rather than a full duration. For instance, the model might generate a caption such as: \textit{``At 00:01, Judy says `Sir, you were going 115 miles per hour.' ''}. In this case, while the start time is explicitly defined, the end time is implicit. Consequently, the parser is configured to output the interval [00:01, Unk].

\noindent \textbf{Scenario 2: Segmentation Granularity Mismatch.}
The second scenario arises when the segmentation of the generated text differs from the ground truth annotation, typically when the model aggregates multiple short sentences into a single block. Consider the following example:
\begin{itemize}
    \item \textbf{Ground Truth:} 
    \begin{itemize}
        \item Sentence A (\textit{``Sir, you were going 115 miles per hour.''}): 00:01--00:03
        \item Sentence B (\textit{``I hope you have a good explanation.''}): 00:04--00:05
    \end{itemize}
    \item \textbf{Model Output:} \textit{``00:01-00:05, Judy says `Sir, you were going 115 miles per hour. I hope you have a good explanation.' ''}
\end{itemize}
Here, the model provides a valid global time range (00:01 to 00:05) for the entire speech block, but fails to delineate the specific internal boundary where Sentence A ends and Sentence B starts. In such cases, we attribute the known boundaries to the respective sentences: the first sentence inherits the global start time, resulting in [00:01, Unk], while the last sentence inherits the global end time, resulting in [Unk, 00:05]. This approach allows us to partially credit the model for correctly identifying the boundaries of the dialogue exchange, even if the internal segmentation is coarse.

\section{Detailed Statistics of Our Curated DVD-Bench}
\label{app:data}

\begin{figure}[h]
    \centering
    \begin{subfigure}{0.48\linewidth}
        \centering
        \includegraphics[width=\linewidth]{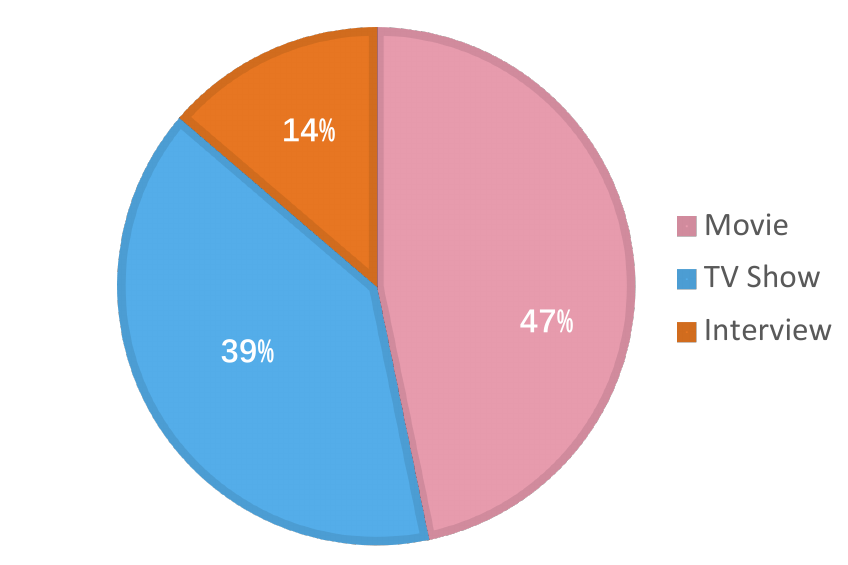}
        \caption{DVD-Bench (En)}
        \label{fig:source_en}
    \end{subfigure}
    \hfill
    \begin{subfigure}{0.48\linewidth}
        \centering
        \includegraphics[width=\linewidth]{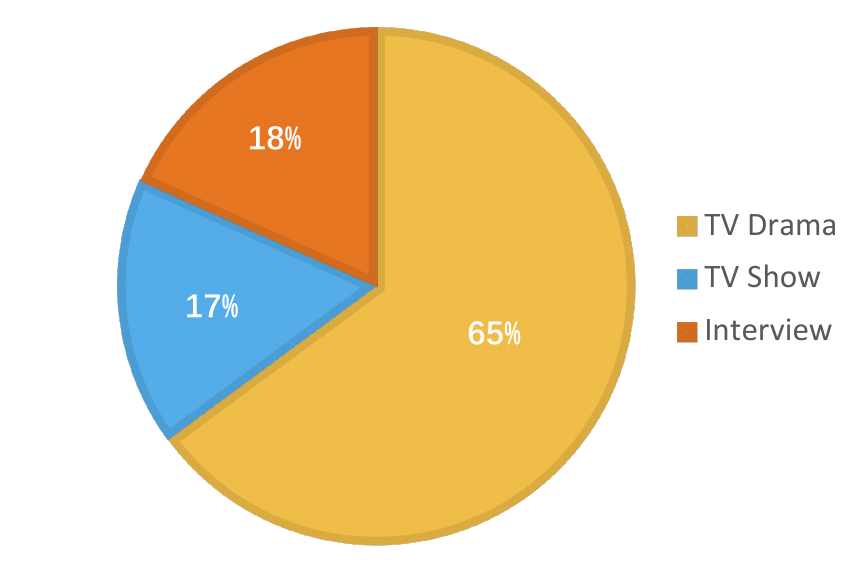}
        \caption{DVD-Bench (Zh)}
        \label{fig:source_zh}
    \end{subfigure}
    \caption{Distribution of video sources in English (En) and Chinese (Zh) subset of our DVD-Bench. The dataset covers a diverse range of genres, including movies, TV dramas, TV shows, and interviews.}
    \label{fig:video_source}
\end{figure}

To ensure a comprehensive assessment of dialogue-centric audio-visual understanding, we compiled a robust evaluation benchmark, DVD-Bench, which consists of 964 English videos and 1,014 Chinese videos. These samples were meticulously curated to cover a wide spectrum of real-world conversational scenarios. The dataset spans from TV dramas, movies, TV shows, and interviews.
The specific distribution of video sources for each language is illustrated in Figure~\ref{fig:video_source}.

Detailed statistical specifications of the benchmark are provided in Table~\ref{tab:benchmarks}. This includes the total number of videos, the average duration, the average number of characters in the video, and the average number of spoken words.

\begin{table}[h]
    \centering
    \caption{Detailed statistics of the DVD-Bench.}
    \small
    \begin{tabular}{lcccc}
    \toprule
    \textbf{Language} &  \textbf{\# Videos} & \textbf{Avg. Duration (s)} & \textbf{Avg. Characters} & \textbf{Avg. Words} \\
    \midrule
     English & 964 & 83.9 & 3.5 & 156.9 \\
    Chinese & 1014 & 94.6 & 5.0 & 330.9 \\
    \bottomrule
    \end{tabular}
    \label{tab:benchmarks}
\end{table}

\section{Training curves of GRPO}
\label{app:training_curves}

\begin{figure}[h]
    \centering
    \begin{subfigure}{0.48\linewidth}
        \centering
        \includegraphics[width=\linewidth]{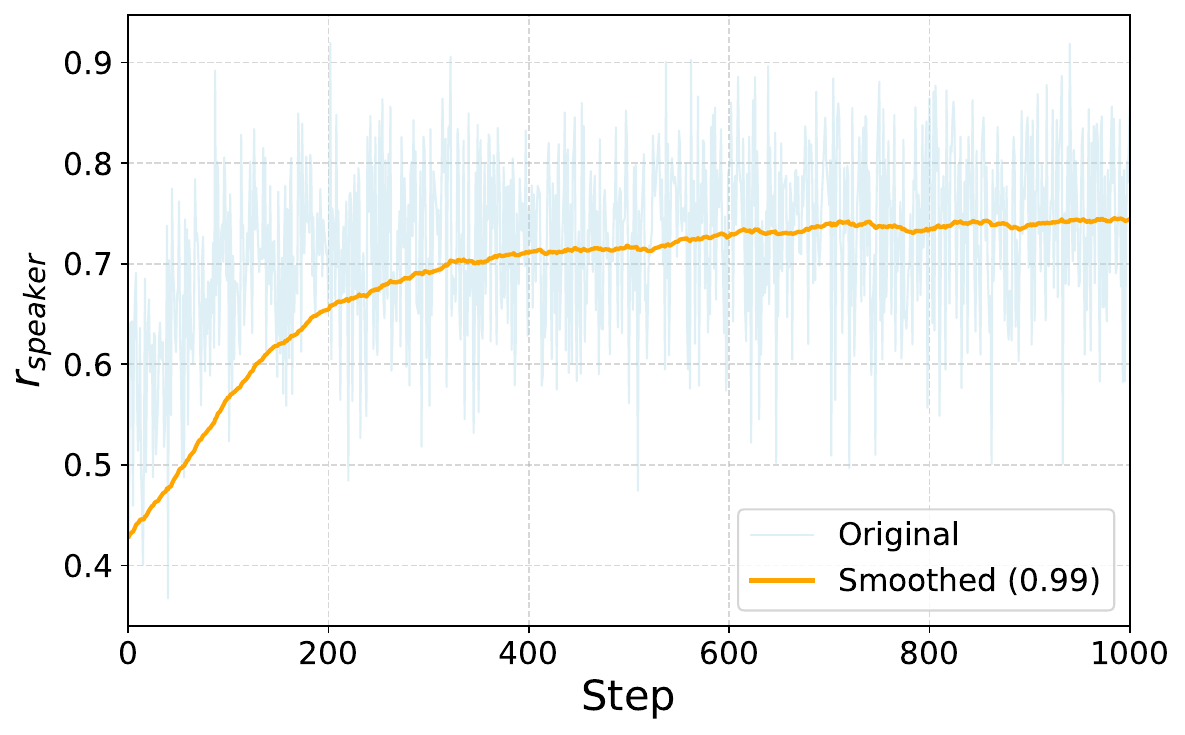}
        \caption{Speaker Attribution Accuracy Reward ($r_{\text{speaker}}$)}
        \label{fig:curve_spk}
    \end{subfigure}
    \hfill
    \begin{subfigure}{0.48\linewidth}
        \centering
        \includegraphics[width=\linewidth]{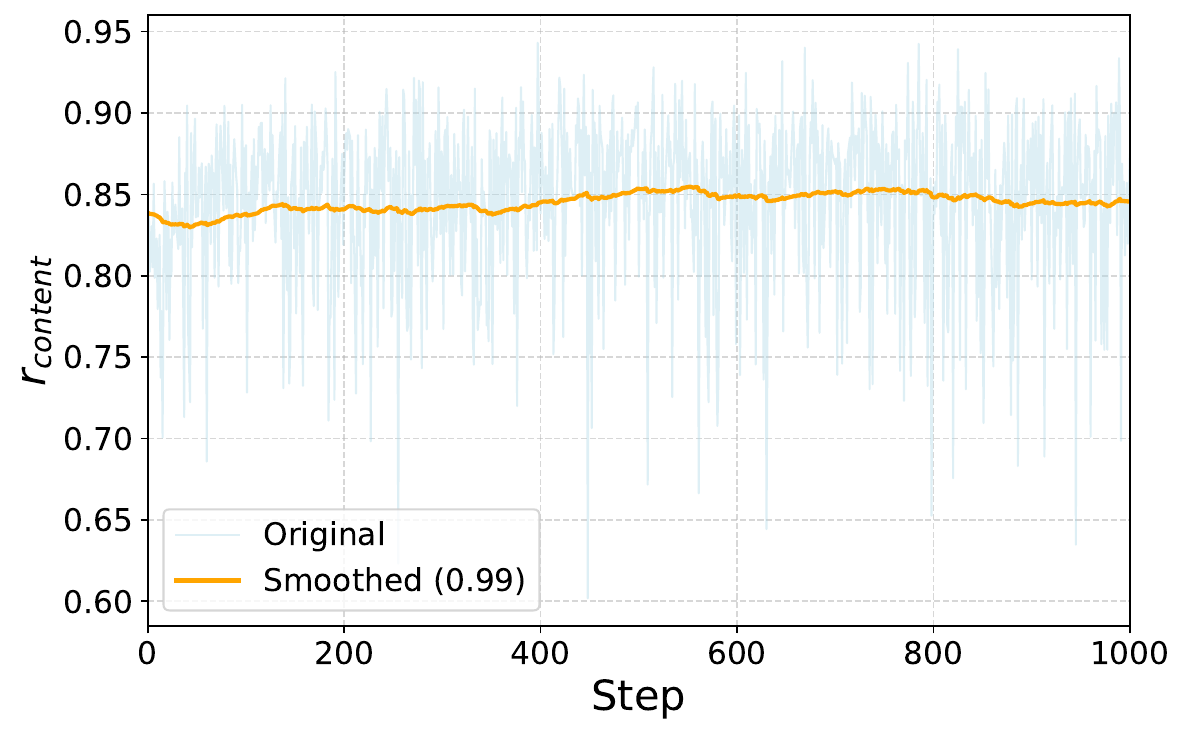}
        \caption{Global Speech Content Reward ($r_{\text{content}}$)}
        \label{fig:curve_wer}
    \end{subfigure}
    
    \vspace{0.3cm} 
    
    \begin{subfigure}{0.48\linewidth}
        \centering
        \includegraphics[width=\linewidth]{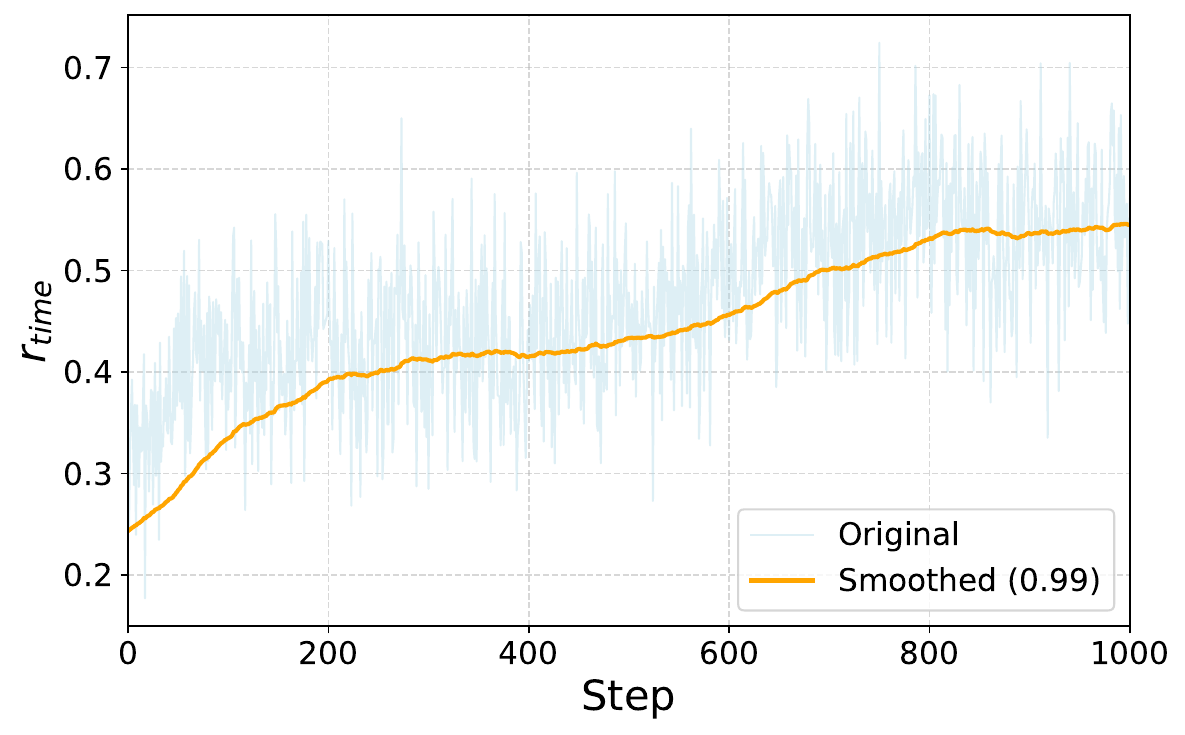}
        \caption{Sentence-level Temporal Reward ($r_{\text{time}}$)}
        \label{fig:curve_time}
    \end{subfigure}
    \hfill
    \begin{subfigure}{0.48\linewidth}
        \centering
        \includegraphics[width=\linewidth]{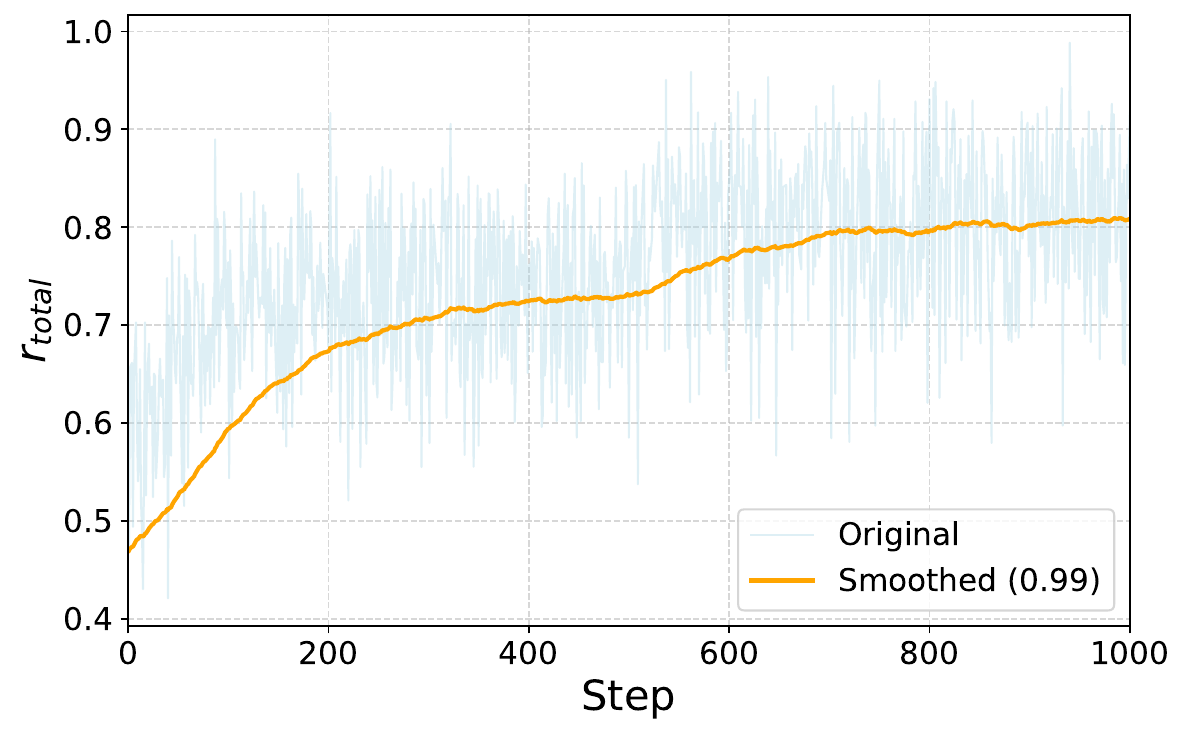}
        \caption{Total Weighted Reward ($r_{\text{total}}$)}
        \label{fig:curve_total}
    \end{subfigure}
    
    \caption{Training curves of D-ORCA during the GRPO stage. The plots demonstrate the progressive improvement of the model across (a) speaker attribution accuracy, (b) speech content fidelity, (c) sentence-level temporal precision, and (d) the overall objective reward function.}
    \label{fig:reward_curves}
\end{figure}

We visualize the training curves of the D-ORCA model during the GRPO stage for reproduction. Figure~\ref{fig:reward_curves} illustrates the trajectories of the three individual reward components ($r_{\text{speaker}}$, $r_{\text{content}}$, $r_{\text{time}}$) as well as the total weighted reward ($r_{\text{total}}$).

\end{document}